\documentclass[runningheads]{llncs}

\usepackage[T1]{fontenc}
\usepackage{graphicx}
\usepackage{booktabs}
\usepackage{multirow}
\usepackage{amsmath,amssymb,amsfonts}
\usepackage{xcolor}
\usepackage{url}
\usepackage{tikz}
\usetikzlibrary{arrows.meta,calc,positioning,shadings}
\usepackage{bbding}

\begin{document}

\title{Ethical LLM-Assisted Research: A Framework for Responsible Delegation, Verification, and Epistemic Value}
\titlerunning{A Framework for Responsible AI Delegation}

\author{Kalin Stoyanov\orcidID{0000-0001-5031-951X}\Envelope
\and Nina Ilieva \orcidID{0009-0008-7146-1986}}
\authorrunning{K. Stoyanov and N. Ilieva}

\institute{Department of Automation, University of Chemical Technology and Metallurgy, Sofia, Bulgaria\\
	\email{kalin.stoyanov@uctm.edu, nilieva@uctm.edu}}

\maketitle

\begin{abstract}
	Large language models (LLMs) increasingly participate in scientific
	research, including literature synthesis, code generation, data
	analysis, and formal reasoning. This paper develops an ethical
	framework in which the legitimacy of LLM-assisted scientific claims
	depends on effective human guidance control, verification, and
	identifiable human responsibility rather than the degree of machine
	participation. The framework distinguishes continuous executor
	provenance $O(g)\in[0,1]$, binary verification $V(g)$, presentation
	status $P(g)$, and the set of responsible humans $R(g)$. Presenting
	a claim as warranted requires $V(g)=1$ and $R(g)\neq\varnothing$,
	independently of provenance. Verification establishes completion
	of the required human-controlled process, while epistemic assessment
	and the decision to present the claim remain distinct human
	judgments. Peer review provides subsequent independent scrutiny
	rather than substituting for author verification. The framework
	also introduces epistemic records to support traceability,
	examination of delegated reasoning, and the identification of
	unresolved questions concerning intellectual provenance.
\end{abstract}


\section{Introduction}
\label{sec:intro}

Large language models (LLMs) and related AI systems are increasingly used in
scientific work for literature synthesis, argument development, coding, data
analysis, and formal reasoning. Their use extends an established feature of
research: cognitive work is distributed across researchers, collaborators,
instruments, software, and institutions. LLMs expand the complexity and scope
of tasks that can be delegated to computational systems.

AI-assisted research raises questions of accountability, transparency,
reproducibility, and trust. Existing guidance assigns responsibility for
scholarly content to human authors, requires disclosure of substantive AI
use, and excludes AI systems from authorship
\cite{ResnikHosseini2025,ICMJE2026,WAME2023,NatureAI2026}.
These rules establish important obligations but leave a further question:
under what conditions may a researcher responsibly present an AI-generated
candidate as a scientifically warranted claim?

The proposed framework separates executor provenance from human verification.
The continuous coordinate $O(g)\in[0,1]$ describes the degree of human--LLM
participation in producing candidate claim $g$. The binary variable
$V(g)\in\{0,1\}$ records completion of the required human-controlled
verification cycle, while $P(g)\in\{0,1\}$ indicates whether the claim is
presented as warranted. The set $R(g)$ identifies the humans who assume
responsibility for that presentation. The central necessary condition is

\begin{equation}
	P(g)=1
	\quad\Longrightarrow\quad
	V(g)=1\land R(g)\neq\varnothing,
	\qquad O(g)\in[0,1].
	\label{eq:intro-condition}
\end{equation}

Provenance is descriptive rather than ethically decisive. The responsible
researcher retains effective guidance control over the delegated process,
ensures completion of the required verification cycle, and assumes
responsibility for presenting the resulting claim.

Verification and epistemic assessment serve distinct functions.
Completion of verification establishes that the required human-controlled
examination has been performed; it does not guarantee truth, originality,
or scientific value. The researcher subsequently evaluates the claim's
evidential support and epistemic significance and decides whether to
present, revise, qualify, or abandon it. Accordingly, $V(g)=1$ does
not imply $P(g)=1$.

The analysis treats LLM-assisted research as \emph{delegated cognition},
drawing on distributed cognition and the ``extended mind'' thesis
\cite{ClarkChalmers1998}. The human researcher remains responsible for
the scientific objective and its warranted presentation, although
operational work may be performed by humans, LLMs, or mixed teams.
Recent developments in mathematics illustrate that such delegation
can extend to research-level reasoning
\cite{Sothanaphan2026,Tao2026MathematicsAI,KlowdenTao2026,JiangEtAl2026}.

The central proposition is:
\begin{quote}
	The legitimacy of delegated scientific reasoning depends on effective
	human guidance control, completed human verification, and identifiable
	human responsibility when a claim enters the scientific record,
	independently of executor composition.
\end{quote}

The paper makes four contributions. First, it introduces a model
combining executor provenance $O(g)$ and binary verification $V(g)$.
Second, it formalizes verification, presentation, and human
responsibility, with guidance control as an additional qualitative
requirement. Third, it distinguishes author verification from
independent peer review. Fourth, it proposes an epistemic record
for documenting delegated reasoning, examining inferential
transitions, and escalating unresolved concerns about
intellectual provenance.

The framework is qualitative. Its possible regulatory implications
and quantitative extensions are identified as directions for
future work.

\section{Background: Delegation and Verification in Scientific Practice}
\label{sec:background}

Scientific research has long relied on cognitive delegation through
collaborators, instruments, software, databases, and formal systems.
AI extends this practice by undertaking increasingly complex reasoning
tasks. Although formal proof assistants such as \emph{Lean} and
\emph{Isabelle} differ fundamentally from generative systems such as
\emph{GPT} and \emph{Gemini}, all illustrate how computational artifacts
can contribute to scientific work under human direction.

The distinction between generating and justifying scientific claims
is well established. Lakatos describes mathematical knowledge as
developing through conjecture, criticism, counterexamples, and
reconstruction \cite{Lakatos1976}. Studies of mathematical practice
likewise distinguish investigative from justificatory activities
\cite{Schlimm2012}. A claim's mode of generation and the grounds
supporting its acceptance are therefore separate matters.

Distributed cognition locates cognitive processes within coordinated
systems of people and artifacts \cite{hutchins1995cognition}.
LLM-generated arguments, code, analyses, and inferences can contribute
to such processes. The distribution of cognitive operations, however,
does not determine the distribution of moral responsibility.
Scientific authorship requires identifiable humans who accept
accountability for the claims they present, as reflected in current
publication guidance \cite{ICMJE2026,WAME2023,NatureAI2026}.

Mitchell \cite{Mitchell2026} criticizes anthropomorphic accounts of
AI behavior that divert attention from human decisions concerning
system design, training, oversight, security, and deployment.
In scientific research, such metaphors can similarly obscure the
responsibility of researchers who determine how AI systems are
used and which generated claims enter the scientific record.

Human verification connects delegated execution to responsible
authorial judgment. It comprises claim-appropriate examination of
methods, evidence, and inferential steps, including correction,
repetition, or revision when required. Effective guidance control
extends throughout this process. Its completion establishes that
the required verification activities have been performed, while
the assessment of scientific validity, significance, and
presentability remains a distinct authorial judgment.

Research on audit practices emphasizes traceability, supporting
evidence, and identifiable accountability
\cite{strathern1997improving,shore2015governing,power1997audit}.
These considerations motivate the separation of executor provenance,
human verification, presentation status, and responsibility developed
in the following sections. They also provide a basis for examining
documented inferential steps and identifying questions that require
further investigation.

\section{Delegated Reasoning and Distributed Cognition}
\label{sec:delegated}

\subsection{Human--Machine Collaboration}
\label{sec:human-machine}

Consider a delegated research arrangement in which a lead researcher
$S$ assigns operational tasks to a team of human executors,

\[
\mathcal{E}_H^{(n)}=\{s_1,\ldots,s_n\},
\qquad n\geq 1.
\]

Drawing on Fischer and Ravizza's account of guidance control
\cite{FischerRavizza1998}, we interpret the role of $S$ as
retaining effective control over the research process and responding
to reasons for approving, correcting, repeating, or rejecting its
outputs. Such control requires substantive involvement in scientific
framing, evaluation of the proposed approach, examination of
evidence, and decisions concerning the resulting claims, while
allowing operational tasks to be delegated.

Researcher $S$ initially defines the scientific problem, its
objectives and constraints, and the criteria by which candidate
results will be evaluated. The executor group implements the
selected approach and produces observations, analyses, intermediate
results, or candidate claims. The researcher subsequently evaluates
their adequacy and may request corrections, repeat operations,
revise the approach, or terminate the investigation.

This process can be represented as an iterative
Plan--Do--Check--Act (PDCA) cycle \cite{MoenNorman2010}.
\emph{Plan} corresponds to scientific framing and selection of an
approach; \emph{Do} to delegated execution; \emph{Check} to
examination of procedures, evidence, and candidate claims; and
\emph{Act} to decisions to accept, correct, repeat, redirect, or
abandon the current line of inquiry. PDCA operationalizes
reasons-responsive guidance control without introducing a separate
condition of moral responsibility.

Human verification is distributed across these iterations.
Deficiencies detected during evaluation can initiate renewed
planning or execution. Completion of the verification cycle
therefore requires an adequate examination of the candidate claim
and the relevant process through which it was obtained.

Scientific significance and epistemic value may inform the
researcher's decisions throughout the process. Their assessment
remains distinct from verification of the claim's evidential and
inferential support. Following verification, $S$ makes the
authorial decision whether to present, revise, further develop,
or abandon the candidate result.

Let $R(g)$ denote the set of identifiable human researchers
assuming responsibility for presenting claim $g$ as warranted.
If $S$ makes that decision, then

\begin{equation}
	S\in R(g),\qquad R(g)\neq\varnothing.
	\label{eq:resp}
\end{equation}

Other participating humans may also belong to $R(g)$.
In this canonical arrangement, operational delegation occurs
within a human team, while guidance control, verification,
authorial judgment, and responsibility remain attributable
to human researchers.

We now consider partial or complete replacement of human
executors by LLMs. The corresponding executor groups are

\[
\mathcal{E}_{HL}^{(k,m)}
=\{s_1,\ldots,s_k,L_1,\ldots,L_m\},
\qquad k,m\geq 1,
\]

and

\[
\mathcal{E}_{L}^{(m)}
=\{L_1,\ldots,L_m\},
\qquad m\geq 1.
\]

Complete replacement concerns only the executor layer:
$S$ remains the human researcher directing the scientific
process and assuming responsibility for the claims presented.

Across all three configurations, $S$ retains control over
problem formulation, selection and revision of the research
approach, examination of evidence, and the final authorial
decision. Machine participation changes the composition of
the executor group while leaving these responsibilities with
the human researcher.

Effective guidance control requires that $S$ can intervene
throughout the process, request additional evidence, examine
inferential steps, correct errors, reconsider assumptions,
and reject unsupported results. In PDCA terms, the researcher
retains meaningful control over \emph{Plan}, \emph{Check},
and \emph{Act}, while human or machine executors may perform
the operational work in \emph{Do}. The ability to return
from \emph{Check} or \emph{Act} to renewed planning and
execution is essential to this iterative control.

Consequently, replacing $\mathcal{E}_H^{(n)}$ with
$\mathcal{E}_{HL}^{(k,m)}$ or $\mathcal{E}_L^{(m)}$
does not alter the responsibility condition in
Eq.~\eqref{eq:resp}. What matters is the effective exercise
of human guidance control and the presence of identifiable
humans who accept responsibility for presenting the claim.
Formal attribution without substantive control cannot
satisfy these obligations.

Figure~\ref{fig:validity-plane} separates executor provenance
from completion of human verification. The responsibility
condition applies throughout the provenance range whenever
a claim is presented as scientifically warranted.

\begin{figure}[htbp]
	\centering
	\begin{tikzpicture}[
		font=\scriptsize,
		>=Latex,
		axis/.style={-Latex,thick},
		boundary/.style={thick,dashed,draw=black!60},
		region/.style={align=center,font=\bfseries\scriptsize}
		]
		\def\W{9.2}
		\def\H{5.6}
		\def\YB{2.8}
		
		\path[fill=blue!7]
		(0,\YB) rectangle (\W,\H);
		\path[fill=gray!8]
		(0,0) rectangle (\W,\YB);
		
		\draw[boundary] (0,\YB) -- (\W,\YB);
		
		\node[region] at (\W*0.50,\H*0.76)
		{Human, human--LLM mixed, or LLM-generated candidate};
		
		\node[region] at (\W*0.50,\H*0.63)
		{Verification completed:\\
			final authorial decision may follow};
		
		\node[region] at (\W*0.50,\H*0.26)
		{Human, human--LLM mixed, or LLM-generated candidate};
		
		\node[region] at (\W*0.50,\H*0.13)
		{Verification incomplete:\\
			further review or revision required};
		
		\draw[axis] (0,0) -- (0,\H);
		\draw[axis] (0,0) -- (\W,0);
		
		\node[anchor=east] at (-0.08,\YB*0.50) {$0$};
		\node[anchor=east] at (-0.08,{(\YB+\H)*0.50}) {$1$};
		
		\node[rotate=90,anchor=south,font=\bfseries]
		at (-0.58,\H*0.50)
		{Verification $V(g)$};
		
		\node[anchor=west,font=\bfseries\scriptsize]
		at (0.2,-0.48)
		{$O(g)$: human execution $\rightarrow$
			mixed execution $\rightarrow$ LLM execution};
	\end{tikzpicture}
	
	\caption{Executor provenance and human verification are
		independent dimensions. Provenance $O(g)\in[0,1]$
		describes the degree of human--LLM participation in
		execution, while $V(g)\in\{0,1\}$ records completion
		of the required human-controlled verification cycle.
		Epistemic assessment, authorial decision, and
		responsibility remain with human researchers.}
	\label{fig:validity-plane}
\end{figure}

\subsection{Formal Framework of Verification and Responsibility}
\label{sec:formal-framework}

Let $\mathcal{C}$ be the set of candidate scientific claims,
$\mathcal{H}$ the participating humans, and $\mathcal{L}$ the
participating LLMs. For each $g\in\mathcal{C}$, let
$\mathcal{E}_g\subseteq\mathcal{H}\cup\mathcal{L}$ denote the
executor group involved in producing the candidate claim.

Executor provenance is represented by
\[
O:\mathcal{C}\longrightarrow[0,1],
\]
where $O(g)=0$ denotes human execution, $O(g)=1$ LLM
execution, and intermediate values indicate mixed
human--LLM participation. Provenance describes executor
composition without assigning scientific validity or
moral responsibility.

Human verification is represented by
\[
V:\mathcal{C}\longrightarrow\{0,1\}.
\]
The condition $V(g)=1$ records completion of the required
human-controlled verification cycle, while $V(g)=0$
indicates that this process remains incomplete.

Verification encompasses claim-appropriate examination
of the scientific approach, execution, evidence, and
inferential steps, together with any necessary corrections,
repetitions, or revisions. Depending on the claim, this
may involve mathematical, empirical, computational,
bibliographic, or interpretive checks. Verification is
therefore an iterative process under human guidance
control rather than a single terminal operation.

Completion of verification establishes that the required
examination has been performed. It does not guarantee
the truth, originality, epistemic value, or
publishability of the candidate claim.

Epistemic assessment may evolve throughout the
verification process. Following its completion,
researcher $S$ forms the final assessment relevant
to presentation,
\[
\widehat{Q}_{\mathrm{epi}}^{\,S}(g),
\]
and decides whether the claim should be presented
as scientifically warranted.

Presentation status is defined by
\[
P:\mathcal{C}\longrightarrow\{0,1\},
\]
where $P(g)=1$ indicates that $g$ is presented as
warranted and $P(g)=0$ that it is not.

Human responsibility is represented by
\[
R:\mathcal{C}\longrightarrow\mathcal{P}(\mathcal{H}),
\]
where $R(g)$ is the set of identifiable humans who
assume responsibility for presenting $g$ as warranted.
By definition, $R(g)\subseteq\mathcal{H}$; an LLM
cannot belong to this set.

The necessary condition for responsible presentation is
\begin{equation}
	\forall g\in\mathcal{C}:\quad
	P(g)=1
	\Longrightarrow
	\bigl(V(g)=1\land R(g)\neq\varnothing\bigr).
	\label{eq:responsible-presentation}
\end{equation}

Equation~\eqref{eq:responsible-presentation}
formalizes the necessary verification and responsibility
conditions. Effective human guidance control remains
an additional qualitative requirement governing the
delegated research process.

In the canonical arrangement, researcher $S$ retains
guidance control and makes the final authorial decision.
If $S$ presents $g$ as warranted, then $S\in R(g)$.
Other participating human researchers may also assume
responsibility.

Equation~\eqref{eq:responsible-presentation} expresses
an invariant under changes in executor composition:
\[
\mathcal{E}_H
\longrightarrow
\mathcal{E}_{HL}
\longrightarrow
\mathcal{E}_L.
\]
These changes affect $\mathcal{E}_g$ and $O(g)$,
while the requirements for human verification and
identifiable responsibility remain unchanged.

The condition is necessary rather than sufficient.
Scientific warrant additionally depends on the
substantive adequacy of the evidence and reasoning,
while responsible publication may involve further
obligations concerning originality, attribution,
disclosure, and research integrity. These obligations
cannot be inferred from executor provenance alone.

\subsection{Human Responsibility as a Moral Commitment}
\label{subsec:moral_commitment}

The responsibility represented by $R(g)$ extends beyond the formal
identification of accountable humans. It entails a personal moral
commitment to the scientific justification of the claim. A researcher
assuming this responsibility accepts an obligation to understand
the essential evidential and inferential chain supporting $g$ and
to be prepared to explain, defend, or reconsider its scientific
grounds when challenged.

This commitment concerns the researcher's accountable judgment
rather than the ability to independently reproduce every technical
operation or rediscover every inferential step. Personal conviction
alone cannot establish scientific warrant. The researcher must
instead have sufficient grounds for accepting the justification,
including the results of appropriate verification and the
examination of material uncertainties.

Where AI-generated reasoning is documented in an epistemic record,
this obligation includes examining significant inferential
transitions and requesting further justification when necessary.
Unresolved questions concerning the logical validity or
intellectual provenance of a contribution may require additional
investigation before the claim is presented as warranted.

Verification and responsibility therefore have complementary
functions: $V(g)$ records completion of the required human-controlled
verification process, whereas $R(g)$ identifies the humans who
personally accept accountability for the resulting scientific
judgment. These obligations apply independently of executor
composition and the degree of machine participation $O(g)$.

\subsection{Author Verification and Peer Review}
\label{sec:peer-review}

Author verification and peer review serve distinct but complementary
functions. Before submitting a claim as scientifically warranted,
authors must complete the required human-controlled verification
process and assume responsibility for its justification, as specified
in Eq.~\eqref{eq:responsible-presentation}. This responsibility includes
the ability to explain, defend, reconsider, or withdraw the claim
in response to scientific criticism.

Peer review provides independent scrutiny of the submitted work.
Reviewers assess the adequacy of its methods, evidence, reasoning,
uncertainty, and reporting against the standards of the discipline
and publication venue. They may identify unsupported inferences,
methodological deficiencies, or insufficient verification and request
clarification, correction, or additional analysis. Responsibility
for performing these revisions remains with the authors, while
editors retain responsibility for editorial decisions.

These roles remain unchanged across human, mixed human--LLM, and
LLM-based executor groups. Scientific claims are assessed against
the same disciplinary standards regardless of executor composition.
However, LLM-assisted work may require additional documentation
to make delegated operations and their verification assessable.
Relevant epistemic records can support such scrutiny by preserving
significant inferential transitions and the evidence used to
evaluate them.

Peer review therefore examines the justification supplied by
responsible authors rather than substituting for their verification.
Its independence provides an additional opportunity to identify
errors, challenge assumptions, and initiate renewed examination
of claims presented as warranted.

\section{Applied Reflection: Responsible Use of AI in Authorship}
\label{sec:applied}

The framework provides practical principles for researchers who
delegate reasoning, analysis, coding, or writing to LLMs. These
principles concern the exercise of human guidance control, the
documentation and verification of delegated work, and the
responsible presentation of scientific claims.

\subsection{Core Principles for Researchers}
\label{sec:core-principles}

Responsible LLM-assisted research follows five principles:

\begin{enumerate}
	
	\item \textbf{Human guidance control.}
	Researchers define the scientific problem, assumptions,
	objectives, and verification criteria. They retain the
	authority and practical ability to evaluate, correct, redirect,
	or reject delegated work. Operational execution may be
	delegated, while scientific direction and authorial
	decisions remain under human control.
	
	\item \textbf{Proportionate traceability.}
	Documentation reflects the epistemic significance of
	delegated operations. Contributions affecting central
	claims, evidence, or inferential steps require greater
	traceability than routine linguistic or presentational
	assistance. Significant reasoning processes should be
	documented during execution to support subsequent
	human examination.
	
	\item \textbf{Claim-appropriate verification.}
	Verification procedures depend on the nature of the
	claim and may include bibliographic, mathematical,
	computational, empirical, or interpretive examination.
	Researchers must be able to request additional
	justification for significant inferential transitions.
	Unresolved questions require further examination
	before verification can be considered complete.
	
	\item \textbf{Human assessment, decision, and
		responsibility.}
	Researchers evaluate the epistemic support and
	significance of candidate claims and decide whether
	they warrant presentation. The condition $P(g)=1$
	requires $V(g)=1$ and $R(g)\neq\varnothing$,
	independently of $O(g)$. Responsible humans
	accept the obligation to explain, defend, or
	reconsider the scientific grounds of the
	claims they present.
	
	\item \textbf{Transparency.}
	Material AI participation must be disclosed according
	to applicable requirements
	\cite{ICMJE2026,WAME2023,NatureAI2026}.
	Disclosure facilitates independent assessment of
	delegated work and its verification. Questions
	concerning the intellectual origin of significant
	contributions should be documented and investigated
	when appropriate.
	
\end{enumerate}

\subsection{Epistemic Assessment and Authorial Decision in Practice}
\label{sec:positive-example}

Consider three candidate claims generated by an LLM.

For $g_1$, the required human-controlled verification cycle is
completed, $V(g_1)=1$, and the researcher assesses its scientific
support positively. The researcher may present the claim as
warranted, setting $P(g_1)=1$ and assuming responsibility
($R(g_1)=\{S\}$ in the single-author case). Alternatively,
the researcher may decide not to present it.

For $g_2$, completed verification, $V(g_2)=1$, establishes
that the original claim lacks sufficient support. It cannot
be presented as warranted in that form. The outcome may
nevertheless motivate a revised claim $g_2'$, subject to
its own verification and epistemic assessment.

For $g_3$, verification is complete, but the final epistemic
assessment is negative, mixed, or qualified. Such an assessment
may still justify reporting the result, provided that its
evidential status, uncertainty, and limitations are accurately
represented and identifiable humans assume responsibility
for its presentation.

These cases distinguish completion of verification from
epistemic assessment and authorial decision. Verification
does not compel acceptance or publication, and the assessment
of a candidate claim may support its rejection, revision,
or qualified presentation. The same distinctions apply
regardless of executor provenance $O(g)$.

\subsection{Failure of Accountable Delegation}
\label{sec:negative-example}

Consider a technically convincing LLM-generated section whose
citations, calculations, or central inferences have not been
examined by the researcher. The required human-controlled
verification cycle remains incomplete, $V(g)=0$. Consequently,
the claims cannot responsibly be presented as scientifically
warranted.

The generation of unverified candidate material is a legitimate
part of scientific exploration. The ethical failure occurs when
such material enters a submitted scientific argument as warranted
without adequate verification and authorial assessment. This
effectively transfers part of the authors' checking obligations
to reviewers, whose role is independent scrutiny rather than
substitution for author verification.

A similar failure arises when a researcher accepts a complex
AI-generated inference without sufficiently examining its
supporting steps. Where an epistemic record exposes an unresolved
inferential gap, responsible presentation requires further
examination. Questions concerning possible unacknowledged
intellectual contributions may additionally require investigation
of information provenance.

These failures concern the exercise of human guidance control
and responsibility rather than the machine origin of the
material. The same principles apply to fully human research
when authors present claims whose evidential or inferential
support they have not adequately examined.

\subsection{Practical Epistemic Record}
\label{sec:record-keeping}

An epistemic record documents significant operations and decisions
throughout delegated scientific work. Its scope should be
proportionate to the epistemic importance of the task. The record
may include:

\begin{enumerate}
	\item \textbf{Provenance:} significant AI contributions to
	claims, analyses, code, evidence processing, or literature
	synthesis, including recorded intermediate reasoning steps;
	
	\item \textbf{Verification:} human control points, sources
	inspected, calculations reproduced, inferential transitions
	examined, and corrective actions taken;
	
	\item \textbf{Assessment, decision, and responsibility:}
	how candidate results were evaluated, why they were presented,
	revised, qualified, or abandoned, and which identifiable
	humans assumed responsibility; and
	
	\item \textbf{Disclosure:} the AI systems used and their
	substantive functions, according to applicable requirements.
\end{enumerate}

For complex AI-generated reasoning, the AI system performs
the delegated task while recording significant intermediate
claims, inferential transitions, and their supporting grounds
as the work proceeds. The record documents the path taken
toward the result rather than reconstructing a plausible
derivation after completion. A recorded sequence may take
the form
\[
g_0\longrightarrow g_1\longrightarrow\cdots
\longrightarrow g_n=g.
\]

Once the result is produced, the lead researcher $S$
examines the recorded sequence in its original direction,
following the development of the argument as if deriving
the result personally. For any selected transition
$g_i\rightarrow g_{i+1}$, $S$ may request additional
intermediate steps, premises, or explanations. The purpose
is human examination of the documented reasoning, while
the AI remains free to conduct the delegated inference
according to its own computational mechanisms.

The examination addresses two distinct questions: whether
each significant transition is logically justified and
whether the intellectual contribution may originate from
an external source. The latter is particularly relevant
when an unpublished scientific contribution previously
supplied to an AI system by another researcher could have
influenced the result.

The record and subsequent explanations should provide
sufficiently specific grounds for $S$ to examine each
significant transition. If $S$ cannot establish its
justification, the unresolved gap warrants further
examination. Where doubts concerning intellectual origin
remain, $S$ may escalate the case to the model provider
or an authorized independent auditor. Such escalation
initiates an investigation rather than establishing
unauthorized appropriation.

Investigation may require access to interaction records,
retrieval mechanisms, shared resources, or other technical
evidence unavailable to $S$, subject to confidentiality
and privacy safeguards. Its purpose is to determine
whether external information could have influenced the
result through the system's available information pathways.
The record supports human verification without claiming
to expose every internal computation of the model. 
Model-generated explanations may not faithfully reflect
the processes underlying their conclusions
\cite{turpin2023language}.

In simplified form, the procedure is
\[
\boxed{\text{delegate}\rightarrow\text{record}\rightarrow
	\text{inspect}\rightarrow\text{verify}\rightarrow
	\text{assess}\rightarrow\text{decide}.}
\]
Inspection and verification may trigger renewed execution,
additional justification, or escalation. The lead researcher
retains guidance control and responsibility for deciding
whether the claim may be presented as warranted, subject
to $V(g)=1$ and $R(g)\neq\varnothing$, independently of $O(g)$.

This approach may also inform future regulatory requirements
for preserving epistemic records and enabling independent
investigation of suspected unauthorized use of intellectual
contributions.

\subsection{Reflexive Application to the Present Article}
\label{sec:reflexive-application}

The disclosure accompanying this article applies the proposed
framework to its own preparation. AI-generated suggestions were
treated as candidate material, while the human authors retained
guidance control over their selection, examination, revision,
integration, and presentation. The authors assume responsibility
for the scientific arguments and claims included in the manuscript.

This application also illustrates the distinction between
disclosure and verification. Reporting AI assistance identifies
aspects of the research process but does not independently
establish the adequacy of the authors' examination. Similarly,
an epistemic record can support traceability and critical
inspection without, by itself, establishing the validity or
independent intellectual origin of the recorded contributions.

Transparency, verification, and personal responsibility thus
perform complementary functions in accountable scientific
authorship.

\section{Future Directions: Quantitative Extension of the Framework}
\label{sec:future}

The proposed framework is qualitative, but several of its
dimensions could support quantitative evaluation. Future work
may develop measures of guidance-control strength, verification
quality, epistemic and scientific value, impact, moral
acceptability, traceability, uncertainty, and clarity of
responsibility. Let $\mathbf{Q}(g)$ collect these assessments.
A research workflow could then be represented by

\[
\mathbf{Z}(g)=
\bigl(O(g),\mathbf{Q}(g),V(g),P(g),R(g)\bigr).
\]

The components of $\mathbf{Q}(g)$ would supplement the
framework's discrete conditions. For example, verification
quality could measure the reliability and depth of the
examination performed, while $V(g)\in\{0,1\}$ would continue
to indicate completion of the required human-controlled
verification cycle.

Quantitative measures could be investigated using epistemic
records, independent audits, replication outcomes, error
rates, and subsequent validation. They could support
comparisons among human, mixed, and LLM-based research
workflows without treating executor provenance as a proxy
for scientific quality or reducing responsibility to an
aggregate score.

Epistemic records also suggest a direction for technical
standardization and future regulation. Requirements for
record preservation, examination of significant inferential
transitions, and independent investigation of substantiated
concerns about intellectual provenance could strengthen
scientific accountability. Their implementation would
require reliable recording mechanisms, appropriate
evidentiary standards, and safeguards for confidential
research information.

Regardless of $O(g)$ and the quantitative assessments in
$\mathbf{Q}(g)$, responsible presentation remains subject
to $V(g)=1$ and $R(g)\neq\varnothing$. Future extensions
should preserve effective human guidance control,
independent epistemic assessment, and identifiable
human responsibility.

\section{Conclusion}
\label{sec:conclusion}

LLMs expand the scope of cognitive work that can be delegated in
scientific research. The proposed framework separates executor
provenance $O(g)$, human verification $V(g)$, presentation status
$P(g)$, and human responsibility $R(g)$. Its central condition
for presenting a claim as scientifically warranted is

\begin{equation}
	P(g)=1
	\quad\Longrightarrow\quad
	V(g)=1\land R(g)\neq\varnothing,
	\qquad O(g)\in[0,1].
	\label{eq:conclusion-condition}
\end{equation}

This condition is independent of executor composition. Human
researchers retain effective guidance control over delegated
work, complete the required verification process, assess the
epistemic grounds of candidate claims, and assume personal
responsibility for their presentation. Verification remains
distinct from epistemic assessment and does not guarantee
truth, originality, or scientific value. Peer review provides
subsequent independent scrutiny rather than replacing
authorial verification.

Epistemic records support accountable delegation by documenting
significant inferential steps during execution and enabling
targeted human examination. Unresolved concerns about the
intellectual origin of a contribution may require escalation
to authorized parties with access to relevant technical evidence.
This procedure also suggests a possible foundation for future
standards and regulatory requirements concerning the traceability
of AI-generated scientific results.

As Mitchell \cite{Mitchell2026} argues, anthropomorphic metaphors
can divert public attention from human decisions toward the AI
system itself. In scientific authorship, such metaphors may
implicitly obscure or displace responsibility for the use of AI
and the claims presented. The proposed framework locates that
responsibility in identifiable human researchers, independently
of the degree of machine participation.

\section*{Acknowledgments}
The authors thank the participants in the discussion
following the presentation of this work at AIMSA 2026.
The discussion motivated the extension of the epistemic
record in Section~\ref{sec:record-keeping}, particularly
the examination of individual inferential transitions
and the escalation of unresolved concerns about
intellectual provenance.

\section*{Data Availability}

No datasets were generated or analyzed in this study.
Data availability is therefore not applicable.

\section*{Generative AI Assistance and Author Responsibility}

OpenAI's ChatGPT was used during the preparation and
revision of this manuscript for language editing,
structural reorganization, critical examination of
arguments, and refinement of formal formulations.
AI-generated suggestions were treated as candidate
material subject to human evaluation and verification.

The human authors retained guidance control over the
conceptual framework, formal definitions, scientific
arguments, and final presentation. They reviewed the
substantive claims and cited sources, determined which
AI-assisted contributions to include, and accept full
responsibility for the content and integrity of the
manuscript.

In accordance with relevant publication guidance
\cite{WAME2023}, AI assistance is disclosed, while
authorship and scientific responsibility remain with
the identifiable human authors.

\begin{credits}
	\subsubsection{\discintname}
	The authors have no competing interests to declare
	that are relevant to the content of this article.
\end{credits}


\begin{thebibliography}{99}


\bibitem{ClarkChalmers1998}
Clark, A., Chalmers, D.J.:
The extended mind.
Analysis \textbf{58}(1), 7--19 (1998).
\doi{10.1093/analys/58.1.7}

\bibitem{FischerRavizza1998}
Fischer, J.M., Ravizza, M.:
Responsibility and Control: A Theory of Moral Responsibility.
Cambridge University Press, Cambridge (1998).
\doi{10.1017/CBO9780511814594}

\bibitem{hutchins1995cognition}
Hutchins, E.:
Cognition in the Wild.
MIT Press, Cambridge, MA (1995)

\bibitem{ICMJE2026}
International Committee of Medical Journal Editors:
Recommendations for the conduct, reporting, editing, and publication of scholarly work in medical journals: Use of artificial intelligence in publishing.
Updated January 2026; accessed August 23, 2026.
\url{https://www.icmje.org/recommendations/browse/artificial-intelligence/} (2026)

\bibitem{JiangEtAl2026}
Jiang, E., Liang, X., Wan, Y., Wan, Y., Li, M., Deng, H., et al.:
From solvers to research: Large language model-driven formal mathematics at the research frontier.
arXiv preprint arXiv:2607.07779 (2026).
\url{https://arxiv.org/abs/2607.07779}

\bibitem{KlowdenTao2026}
Klowden, T., Tao, T.:
Mathematical methods and human thought in the age of AI.
arXiv preprint arXiv:2603.26524 (2026).
\url{https://arxiv.org/abs/2603.26524}

\bibitem{Lakatos1976}
Lakatos, I.:
Proofs and Refutations: The Logic of Mathematical Discovery.
Cambridge University Press (1976)

\bibitem{Mitchell2026}
M. Mitchell,
``Misleading Metaphors and Real Risks: What To Fear from AI Agents and How to Reclaim Our Human Agency,''
\textit{AI: A Guide for Thinking Humans}, Substack, September 10, 2026.
Available: \url{https://aiguide.substack.com/p/misleading-metaphors-and-real-risks}

\bibitem{MoenNorman2010}
R.~D. Moen and C.~L. Norman,
``Circling Back: Clearing Up Myths About the Deming Cycle and Seeing How It Keeps Evolving,''
\emph{Quality Progress}, vol.~43, no.~11, pp.~22--28, 2010.

\bibitem{NatureAI2026}
Nature Methods:
Using AI responsibly in scientific publishing.
Nature Methods \textbf{23}, 271 (2026).
\doi{10.1038/s41592-026-03020-1}

\bibitem{power1997audit}
Power, M.:
The Audit Society: Rituals of Verification.
Oxford University Press (1997)

\bibitem{ResnikHosseini2025}
Resnik, D.B., Hosseini, M.:
The ethics of using artificial intelligence in scientific research: New guidance needed for a new tool.
AI and Ethics \textbf{5}(2), 1499--1521 (2025).
\doi{10.1007/s43681-024-00493-8}

\bibitem{Schlimm2012}
Schlimm, D.:
Mathematical concepts and investigative practice.
In: Feest, U., Steinle, F. (eds.) Scientific Concepts and Investigative Practice,
pp. 127--147.
De Gruyter, Berlin/Boston (2012).
\doi{10.1515/9783110253610.127}

\bibitem{shore2015governing}
Shore, C., Wright, S.:
Governing by numbers: Audit culture, rankings and the new world order.
Social Anthropology / Anthropologie Sociale \textbf{23}(1), 22--28 (2015).
\doi{10.1111/1469-8676.12098}

\bibitem{strathern1997improving}
Strathern, M.:
Improving ratings: Audit in the British university system.
European Review \textbf{5}(3), 305--321 (1997).
\doi{10.1002/(SICI)1234-981X(199707)5:3<305::AID-EURO184>3.0.CO;2-4}

\bibitem{Sothanaphan2026}
Sothanaphan, N.:
Resolution of Erdős Problem \#728: a writeup of Aristotle's Lean proof.
arXiv preprint arXiv:2601.07421 (2026).
\url{https://arxiv.org/abs/2601.07421}

\bibitem{Tao2026MathematicsAI}
Tao, T.:
Mathematics in the age of AI.
arXiv preprint arXiv:2608.16753 (2026).
Essay based on a public lecture delivered at the 2026 International Congress of Mathematicians.
\url{https://arxiv.org/abs/2608.16753}

\bibitem{turpin2023language}
Turpin, M., Michael, J., Perez, E., Bowman, S.R.:
Language models don't always say what they think:
Unfaithful explanations in chain-of-thought prompting.
In: Advances in Neural Information Processing Systems,
vol. 36, pp. 74952--74965 (2023).


\bibitem{WAME2023}
World Association of Medical Editors:
Recommendations on chatbots and generative artificial intelligence in relation to scholarly publication.
Accessed August 23, 2026.
\url{https://wame.org/pdf/Chatbots-Generative-AI-and-Scholarly-Manuscripts.pdf} (2023)

\end{thebibliography}
\end{document}